\pdfoutput=1

\documentclass[11pt]{article}

\usepackage[]{emnlp2021}

\usepackage{times}
\usepackage{latexsym}

\usepackage[T1]{fontenc}

\usepackage[utf8]{inputenc}

\usepackage{microtype}

\usepackage{amsmath, amssymb, amsthm}
\usepackage{booktabs}
\usepackage{graphicx}
\usepackage{makecell}
\usepackage{hhline}
\usepackage[caption=false]{subfig}

\usepackage{xcolor}

\usepackage{pgfplotstable}
\usepackage{pgfplots}
\pgfplotsset{width=0.47\textwidth,compat=1.9, height=1.6in}

\usepackage{url}
\usepackage[normalem]{ulem}

\definecolor{lblue}{HTML}{DAE8FC}
\definecolor{dblue}{HTML}{6C8EBF}
\definecolor{dgreen}{HTML}{33A02C}
\definecolor{dred}{HTML}{E31A1C}
\newcommand{\dblue}[1]{{\color{dblue} #1}}
\newcommand{\dgreen}[1]{{\color{dgreen} #1}}
\newcommand{\dred}[1]{{\color{dred} #1}}

\title{Towards Generating Citation Sentences \\ for Multiple References with Intent Control }

\author{Jia-Yan Wu\footnotemark[1]\quad Alexander Te-Wei Shieh\footnotemark[1]\quad Shih-Ju Hsu\footnotemark[1]\quad Yun-Nung Chen\\
National Taiwan University, Taipei, Taiwan \\
  {\tt \{b06701216,b05401009,b07902076\}@ntu.edu.tw\quad y.v.chen@ieee.org } \\}

\begin{document}
\maketitle
\setcounter{footnote}{0}
\renewcommand*{\thefootnote}{\fnsymbol{footnote}}
\footnotetext[1]{These authors contributed equally to this work.}
\begin{abstract}

Machine-generated citation sentences can aid automated scientific literature review and assist article writing. Current methods in generating citation text were limited to single citation generation using the citing document and a cited document as input. However, in real-world situations, writers often summarize several studies in one sentence or discuss relevant information across the entire paragraph. In addition, multiple citation intents have been previously identified, implying that writers may need control over the intents of generated sentences to cover different scenarios. Therefore, this work focuses on generating multiple citations and releasing a newly collected dataset named CiteMI to drive the future research. We first build a novel generation model with the Fusion-in-Decoder approach to cope with multiple long inputs. Second, we incorporate the predicted citation intents into training for intent control. The experiments demonstrate that the proposed approaches provide much more comprehensive features for generating citation sentences.

\end{abstract}

\section{Introduction}

Every month, more than ten thousand papers are submitted to platforms such as arXiv \cite{mckiernan2000arxiv}. The amount of scientific literature have grown enormously in recent years \cite{ginsparg2011arxiv}, so as the amount of citations needed to compose a new publication nowadays. Hence, literature review has become time-consuming, and increasing effort is needed to write them into the ``related work" section. The development of automatic citation text generation system can relieve scientific researcher's burden on tracking and editing citations. 
Automatic summarization already accelerated this process, but systems that can fully automate citation text generation are yet to be explored. Also, different from summarization systems, which typically involve one source document at a time, citation text generation systems need to consider both the citing and the cited documents.

However, current citation text generation systems have several limitations. They focused on generating a single sentence for a single citation, and the intent of citation were not controllable \cite{xing-etal-2020-automatic, luu2020citation}. Previous work demonstrated that citations can be classified into several intentions, such as background information, method or result comparison \citet{cohan-etal-2019-structural}. They are crucial for studying citation behavior and evaluating the impact of a certain publication \cite{SMALL2018461}. As for authors, they often need citation sentences with different intent for the same cited document depending on the current context. 

\begin{figure}[t!]
    \centering
    \includegraphics[width=1.0\linewidth]{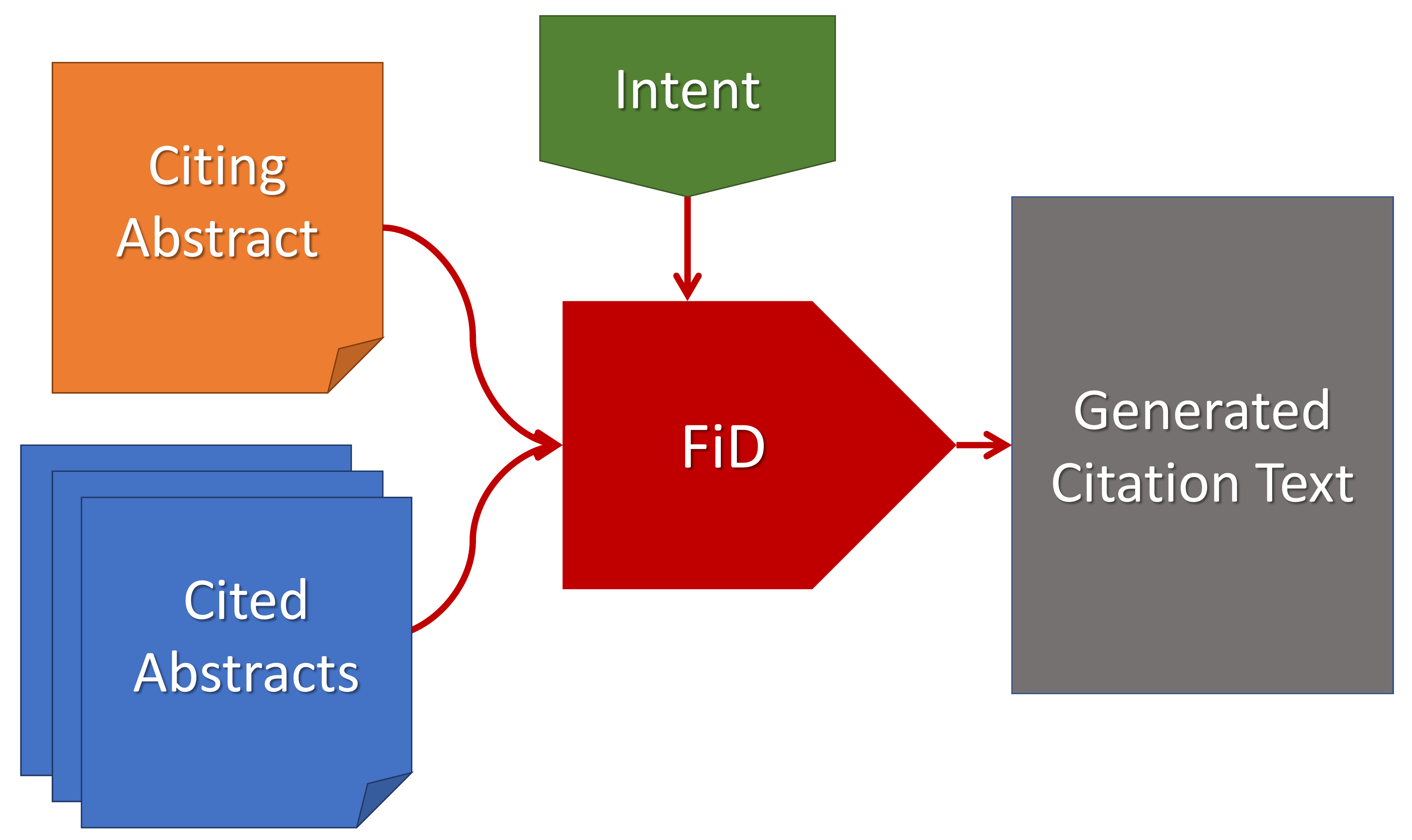}
    \caption{Overview of the CiteMI task using FiD as the generation model.}
    \label{fig:citemi}
\end{figure}

This work proposes methods that address these two problems. First, we collect a new dataset that supports multiple cited documents as its input. Second, we train a model to predict citation intents for previous single citation generation data and our newly collected corpus containing examples with multiple citations. The predicted citation is incorporated into the input of citation sentence generation, providing control over intents in generation. Third, we use the Fusion-in-Decoder (FiD) model \cite{izacard-grave-2021-leveraging} for citation text generation. The FiD model was originally used in generative open domain question answering, but we found it useful for generating sentences as well. We choose FiD for its capability of reading long inputs from multiple documents while benefiting from the prowess of large pre-trained contextual language models such as T5 \cite{JMLR:v21:20-074} for generation.

Our contributions are three-fold: (i) we identify the problem of multiple cited documents and intent control in the citation sentence generation task; (ii) we collect a new dataset, named \textbf{CiteMI} (stands for \textbf{Cite} \textbf{M}ultipe References with \textbf{I}ntent Control), that addresses two problems discussed above; and (iii) we leverage the Fusion-in-Decoder approach to build a model that can serve the above features. The entire workflow is depicted in  Figure~\ref{fig:citemi}. We further evaluate our model with enriched features and discuss future directions on improving this general citation text generation task.

\begin{table*}[t!]
  \small
  \begin{center}
  \def\arraystretch{1.5}
  \begin{tabular}{|p{0.95\textwidth}|}
  \hline
  {\bf Source:} \newline
   \dblue{(Intent)}
   Background \newline
   \dblue{(Citing paper's abstract)}\newline
   This paper presents a neural network system where we participate in the first task of SemEval-2020 shared task 7 \"Assessing the Funniness of Edited News Headlines\". Our target is to create to neural network model that can predict the funniness of edited headlines. We build our model using a combination of LSTM and TF-IDF, then a feed-forward neural network. The system manages to slightly improve RSME scores regarding our mean score baseline.

   \dblue{(Cited paper1's abstract)}
   ... The main challenges for irony generation are the lack of large-scale irony dataset and difficulties in modeling the ironic pattern. In this work, we first systematically define irony generation based on style transfer task. To address the lack of data, we make use of twitter and build a large-scale dataset. We also design a combination of rewards for reinforcement learning to control the generation of ironic sentences... 
   
   \dblue{(Cited paper2's abstract)}
   We introduce the Self-Annotated Reddit Corpus (SARC), a large corpus for sarcasm research and for training and evaluating systems for sarcasm detection. The corpus has 1.3 million sarcastic statements ... We evaluate the corpus for accuracy, construct benchmarks for sarcasm detection, and evaluate baseline methods. \\

  \hhline{|=|}
  {\bf Target (\dgreen{ground truth}):}\newline <B1> collects regular and sarcastic Amazon product reviews to identify sarcasm on a sentence level or for a specific phrase. <B2> introduce the Self-Annotated Reddit Corpus (SARC) which is a corpus that has 1.3 million sarcastic statements for training and evaluating systems for sarcasm detection. \\
  \hline
  {\bf FiD output (without intent):}\newline <B1> used a crowdsourced dataset to generate a sarcastic sentence. \\ 
  \hline
  {\bf FiD output (\dred{with intent}):}\newline <B1> used a crowdsourced corpus for sarcasm detection. <B2> used an unsupervised corpus to generate sarkastic sentences.\\ 
  \hline
  \end{tabular}
  \end{center}
  \caption{An example the CiteMI task: given the intent, citing paper's abstract and cited papers' abstracts, the model needs to output the corresponding citation text. We showed the actual generation result of the FiD models with and without intent. }
  \label{tab:intro}
\end{table*}
\begin{table*}[t!]
\centering
\begin{tabular}{lcccccc}
  \toprule
  Model & BLEU & ROUGE-1 & ROUGE-2 & ROUGE-L & BLEURT & Meteor \\
  \midrule
  Retrieval (oracle)        & 7.13 & 10.63 & 1.17    & 7.92    & -0.76 & 4.31 \\
  Retrieval (baseline)           & 4.11 & 7.34 & 0.63 & 5.38 & -1.42 & 3.42\\
  \midrule
  CiteMI (FiD, without intent)   & 15.94 & 22.96 & 3.05    & 15.76    & -0.86 & 11.15\\
  \textbf{CiteMI (FiD, with intent)}      & \textbf{17.13} & \textbf{23.62} & \textbf{3.37}    & \textbf{16.06}    & \textbf{-0.73} & \textbf{11.67} \\
  \bottomrule
\end{tabular}
\caption[Caption]{Automatic evaluation results of the proposed CiteMI task.}
\label{tab:automatic_results}
\end{table*}

\section{Related Work}

\subsection{Citation Text Generation}

\citet{xing-etal-2020-automatic} defined citations into two types: explicit and implicit citations, meaning citations with or without the author's direct reference respectively. They built a citation text generation dataset based on the ACL Anthology Network corpus \cite{10.1007/s10579-012-9211-2,TACL1266}. In addition, they manually labeled 1,000 citation texts as either explicit or implicit, and used this dataset to train a BERT-based \cite{devlin2019bert} model that extracts more implicit citations. 
Then they used a PG-net \cite{see-etal-2017-get} with a cross attention model for generating the final citation text, given the citing paper's context and the cited paper's abstract. In the data, they used the ``[refer]'' token to indicate explicit citation and the ``[otherrefer]'' token to replace references other than the one in the input.

Followingly, the work by \citet{luu2020citation} also investigated citation text generation given one source document and one cited document. They tried using combinations of the abstract, introduction from the source and cited documents as input. Their training set was based on the computer science articles collected by the S2-GORC dataset \cite{lo-etal-2020-s2orc}, which is significantly larger than the prior work. Instead of PG-net, they used GPT-2 \cite{radford2019language} as their generation model. Their results showed better automatic evaluation scores for the retrieval baseline, and the generation models performed better in human evaluation.


\subsection{Citation Intents}

The work by \citet{cohan-etal-2019-structural} produced the SciCite dataset, which provided 11,020 instances of citations with crowdsourced intents. 
The intents are categorized into three types: ``Background'', ``Method'' and ``Result comparison''.
Compared to earlier work in the ACL-ARC dataset \cite{TACL1266}, the SciCite dataset has fewer categories but comes from more diverse domains and includes much more instances.
They leveraged section title prediction and citation worthiness prediction for improving the intention prediction task in a multi-task setting, showing that three tasks have correlation so that they can benefit each other. Another recent work on citation intent classification asked the authors to classify their own reasons for citation \cite{10.1145/3383583.3398617} and finally form a dataset containing six types of of citations with 11,233 instances.


\subsection{Extreme Summarization}

A similar task related to citation text generation is extreme summarization. The recent study by \citet{cachola-etal-2020-tldr} introduced the SciTLDR dataset. TLDRs are short sentence with only 15-25 words that capture the key aspects of papers. These TLDR sentences resembled citations that provide background information, but are not conditioned on the citing document. They proposed the CATTS approach that utilizes control codes \cite{keskar2019ctrl,he2020ctrlsum} to generate either the title or the TLDR sentence given the abstract, introduction and conclusion of a paper. 


\section{Task Description and Data}

The input of the CiteMI task include: the citing document $A$'s title and abstract, several consecutively cited documents $Bi$'s titles and abstracts and the intent $I_{(A,Bi)}$ for their citation, respectively. The output of this task is the citation sentences of the corresponding cited documents. We define consecutively citation as two explicit citation sentences with only one sentence sentence without explicit citations in-between. The citation sentences were extracted using handcrafted regular expressions.

We obtained the multiple reference citation sentences from the online proceedings of ACL anthology published during 2015-2020. In addition, we added single reference citation examples from  \citet{xing-etal-2020-automatic}, which includes  papers until 2014 from the latest ACL Anthology Network release \cite{10.1007/s10579-012-9211-2}, and the SciCite intent corpus \cite{cohan-etal-2019-structural} to our training set. We then crawled the abstracts of the papers in SciCite, since they were not included in the original release. 

Other than the examples from SciCite, which intent labels were provided, we used a BERT-based \cite{devlin2019bert} model trained on SciCite to predict citation intents using the target citation text for the rest of the CiteMI dataset. The BERT-based model used in this step obtained 84\% testing accuracy on the SciCite dataset. Currently, the CiteMI dataset has 56,955 examples with single citation and 4,189 examples with multiple citations. An example of CiteMI's input and output is shown in Table~\ref{tab:intro}.

\section{Methods and Experiments}

\subsection{Retrieval Baseline}

We first made an oracle retrieval model that returns the most semantically similar sentence from the cited document compared to the ground truth target. The mean of token embeddings from the BERT model \cite{devlin2019bert} was used to represent each sentence in the cited paper's abstract. The sentence  that has the highest cosine similarity compared with the ground truth citation text was returned. Another baseline model was made with the same method but sentences were compared with the citing paper's abstract instead of the target.

\subsection{Generative Models}
Our generative model augmented the T5 \cite{JMLR:v21:20-074} model with the Fusion-in-Decoder (FiD) architecture \cite{izacard-grave-2021-leveraging}. For the nth encoder block in FiD, we encode the citing document $A$'s abstract, together with the code ``<$B_n$>'' followed by the nth cited document $B_n$'s title and abstract. Additionally, we prepended ``<$I_{(A,B_n)}$>'', where $I_{(A,Bn)}$ is either ``background'', ``method'', ``supportive'' or ``not\_supportive'', to the abstract of $A$ to specify the current citation intent. Efficacy of controlling general summarization was discussed in \citet{fan-etal-2018-controllable} and \citet{he2020ctrlsum}. The decoder then performs attention over the encoder blocks that represented citing and the cited papers. Note that the ``supportive'' and ``not\_supportive'' labels are additional labels provided by \citet{cohan-etal-2019-structural} for the ``Result comparison'' intent. As for the target citation text, we replaced the nth citation with <$B_n$>.


By leveraging the FiD architecture, we can include longer data, i.e., more cited documents, to the input. The computation time of our model grows linearly with the number of cited documents, instead of quadratically when using a single encoder block. This is beneficial for scaling to examples with multiple citations in the CiteMI task. For all generation experiments, we used 80\% of the CiteMI data for training, 10\% for validation and 10\% for testing.

\section{Results and Analysis}

In addition to conventional automatic evaluation scores BLEU \cite{papineni-etal-2002-bleu} for machine translation and ROUGE \cite{lin-2004-rouge} for summarization, we also evaluate the generated citation sentences using Meteor \cite{denkowski-lavie-2014-meteor} and BLEURT \cite{sellam-etal-2020-bleurt}. The BLEURT score currently achieves state-of-the-art agreement with human judgement on machine translation benchmarks. Our automatic evaluation results were presented in Table~\ref{tab:automatic_results}. Apparently, adding intent labels further boosted FiD's performance on the CiteMI task. The improvement was unanimous across all evaluation metrics that we evaluated, suggesting a role for explicitly specifying citation intent in generation. 

We also performed a round-trip intent prediction on the generated citation text. When the intent was specified, the generated citation text achieved a micro-average accuracy of 86.39\%. On the other hand, when the intent was not given the accuracy decreased to 70.14\%. These results indicated that explicitly specifying the intent could effectively change the intent expressed by the generation model. Also, it showed that providing only the abstracts without intent could still reveal some clues about the underlying intent.


\section{Future Work}

We suggest further augmentation of the CiteMI data in the near future. Our work here only included papers related to computer science and natural language processing. However, there are more publications emerging in fields such as medicine. Further inclusion of multi-domain data is needed to build a more comprehensive model. Also, since some of the citation intents in our work were predicted, more accurately labeled citation intent should be added. 

On the other hand, since an abstractive model was used in our study, pitfalls for evaluating such systems could occur. For instance, hallucinations, or contents that are not faithful or factual to the input could appear \cite{maynez-etal-2020-faithfulness}. Scientific facts should be presented accurately, methods to examine the correctness of the generated results should therefore be included. Moreover, there are increasing research results regarding to citation recommendation \cite{nogueira2020navigationbased, giosa2020what2cite, ALI2020113790, khadka2020capturing, ali2021overview}. Pipelines combining citation recommendation and  generation can make the workflow of reviewing scientific literature more efficient.

\section{Conclusion}

In conclusion, we addressed the problem of multiple consecutively cited documents in citation text generation by building a new dataset named \textbf{CiteMI}, which contains both single and multiple citation examples in the computer science domain. Also, we added predicted intention labels to provide intent control over the generated content. Although dealing with long inputs containing multiple abstracts could be difficult with Transformer-based models, we demonstrated the feasibility of building a model that can realize both features using the Fusion-in-Decoder approach. Empirically, we observed that adding intent labels improved the generation model's performance on various automatic evaluation metrics. We expect future work on improving the accuracy of the intent labels and increasing the amount of examples with multiple citations can further enhance the overall quality of the generated citation text.

\bibliography{anthology,custom}

\begin{thebibliography}{29}
\expandafter\ifx\csname natexlab\endcsname\relax\def\natexlab#1{#1}\fi

\bibitem[{Ali et~al.(2020)Ali, Kefalas, Muhammad, Ali, and
  Imran}]{ALI2020113790}
Zafar Ali, Pavlos Kefalas, Khan Muhammad, Bahadar Ali, and Muhammad Imran.
  2020.
\newblock \href {https://doi.org/https://doi.org/10.1016/j.eswa.2020.113790}
  {Deep learning in citation recommendation models survey}.
\newblock \emph{Expert Systems with Applications}, 162:113790.

\bibitem[{Ali et~al.(2021)Ali, Ullah, Khan, Jan, and
  Muhammad}]{ali2021overview}
Zafar Ali, Irfan Ullah, Amin Khan, Asim~Ullah Jan, and Khan Muhammad. 2021.
\newblock An overview and evaluation of citation recommendation models.
\newblock \emph{Scientometrics}, pages 1--37.

\bibitem[{Cachola et~al.(2020)Cachola, Lo, Cohan, and
  Weld}]{cachola-etal-2020-tldr}
Isabel Cachola, Kyle Lo, Arman Cohan, and Daniel Weld. 2020.
\newblock \href {https://doi.org/10.18653/v1/2020.findings-emnlp.428} {{TLDR}:
  Extreme summarization of scientific documents}.
\newblock In \emph{Findings of the Association for Computational Linguistics:
  EMNLP 2020}, pages 4766--4777, Online. Association for Computational
  Linguistics.

\bibitem[{Cohan et~al.(2019)Cohan, Ammar, van Zuylen, and
  Cady}]{cohan-etal-2019-structural}
Arman Cohan, Waleed Ammar, Madeleine van Zuylen, and Field Cady. 2019.
\newblock \href {https://doi.org/10.18653/v1/N19-1361} {Structural scaffolds
  for citation intent classification in scientific publications}.
\newblock In \emph{Proceedings of the 2019 Conference of the North {A}merican
  Chapter of the Association for Computational Linguistics: Human Language
  Technologies, Volume 1 (Long and Short Papers)}, pages 3586--3596,
  Minneapolis, Minnesota. Association for Computational Linguistics.

\bibitem[{Denkowski and Lavie(2014)}]{denkowski-lavie-2014-meteor}
Michael Denkowski and Alon Lavie. 2014.
\newblock \href {https://doi.org/10.3115/v1/W14-3348} {Meteor universal:
  Language specific translation evaluation for any target language}.
\newblock In \emph{Proceedings of the Ninth Workshop on Statistical Machine
  Translation}, pages 376--380, Baltimore, Maryland, USA. Association for
  Computational Linguistics.

\bibitem[{Devlin et~al.(2019)Devlin, Chang, Lee, and
  Toutanova}]{devlin2019bert}
Jacob Devlin, Ming-Wei Chang, Kenton Lee, and Kristina Toutanova. 2019.
\newblock \href {http://arxiv.org/abs/1810.04805} {Bert: Pre-training of deep
  bidirectional transformers for language understanding}.

\bibitem[{Fan et~al.(2018)Fan, Grangier, and Auli}]{fan-etal-2018-controllable}
Angela Fan, David Grangier, and Michael Auli. 2018.
\newblock \href {https://doi.org/10.18653/v1/W18-2706} {Controllable
  abstractive summarization}.
\newblock In \emph{Proceedings of the 2nd Workshop on Neural Machine
  Translation and Generation}, pages 45--54, Melbourne, Australia. Association
  for Computational Linguistics.

\bibitem[{Ginsparg(2011)}]{ginsparg2011arxiv}
Paul Ginsparg. 2011.
\newblock Arxiv at 20.
\newblock \emph{Nature}, 476(7359):145--147.

\bibitem[{Giosa and Di~Caro(2020)}]{giosa2020what2cite}
Davide Giosa and Luigi Di~Caro. 2020.
\newblock What2cite: Unveiling topics and citations dependencies for scientific
  literature exploration and recommendation.
\newblock In \emph{International Conference on Knowledge Engineering and
  Knowledge Management}, pages 147--157. Springer.

\bibitem[{He et~al.(2020)He, Kryściński, McCann, Rajani, and
  Xiong}]{he2020ctrlsum}
Junxian He, Wojciech Kryściński, Bryan McCann, Nazneen Rajani, and Caiming
  Xiong. 2020.
\newblock \href {http://arxiv.org/abs/2012.04281} {Ctrlsum: Towards generic
  controllable text summarization}.

\bibitem[{Izacard and Grave(2021)}]{izacard-grave-2021-leveraging}
Gautier Izacard and Edouard Grave. 2021.
\newblock \href {https://www.aclweb.org/anthology/2021.eacl-main.74}
  {Leveraging passage retrieval with generative models for open domain question
  answering}.
\newblock In \emph{Proceedings of the 16th Conference of the European Chapter
  of the Association for Computational Linguistics: Main Volume}, pages
  874--880, Online. Association for Computational Linguistics.

\bibitem[{Jurgens et~al.(2018)Jurgens, Kumar, Hoover, McFarland, and
  Jurafsky}]{TACL1266}
David Jurgens, Srijan Kumar, Raine Hoover, Dan McFarland, and Dan Jurafsky.
  2018.
\newblock \href {https://transacl.org/ojs/index.php/tacl/article/view/1266}
  {Measuring the evolution of a scientific field through citation frames}.
\newblock \emph{Transactions of the Association for Computational Linguistics},
  6(0):391--406.

\bibitem[{Keskar et~al.(2019)Keskar, McCann, Varshney, Xiong, and
  Socher}]{keskar2019ctrl}
Nitish~Shirish Keskar, Bryan McCann, Lav~R. Varshney, Caiming Xiong, and
  Richard Socher. 2019.
\newblock \href {http://arxiv.org/abs/1909.05858} {Ctrl: A conditional
  transformer language model for controllable generation}.

\bibitem[{Khadka(2020)}]{khadka2020capturing}
Anita Khadka. 2020.
\newblock \emph{Capturing and Exploiting Citation Knowledge for the
  Recommendation of Scientific Publications}.
\newblock Ph.D. thesis, The Open University.

\bibitem[{Lin(2004)}]{lin-2004-rouge}
Chin-Yew Lin. 2004.
\newblock \href {https://www.aclweb.org/anthology/W04-1013} {{ROUGE}: A package
  for automatic evaluation of summaries}.
\newblock In \emph{Text Summarization Branches Out}, pages 74--81, Barcelona,
  Spain. Association for Computational Linguistics.

\bibitem[{Lo et~al.(2020)Lo, Wang, Neumann, Kinney, and
  Weld}]{lo-etal-2020-s2orc}
Kyle Lo, Lucy~Lu Wang, Mark Neumann, Rodney Kinney, and Daniel Weld. 2020.
\newblock \href {https://doi.org/10.18653/v1/2020.acl-main.447} {{S}2{ORC}: The
  semantic scholar open research corpus}.
\newblock In \emph{Proceedings of the 58th Annual Meeting of the Association
  for Computational Linguistics}, pages 4969--4983, Online. Association for
  Computational Linguistics.

\bibitem[{Luu et~al.(2020)Luu, Koncel-Kedziorski, Lo, Cachola, and
  Smith}]{luu2020citation}
Kelvin Luu, Rik Koncel-Kedziorski, Kyle Lo, Isabel Cachola, and Noah~A. Smith.
  2020.
\newblock \href {http://arxiv.org/abs/2002.00317} {Citation text generation}.

\bibitem[{Maynez et~al.(2020)Maynez, Narayan, Bohnet, and
  McDonald}]{maynez-etal-2020-faithfulness}
Joshua Maynez, Shashi Narayan, Bernd Bohnet, and Ryan McDonald. 2020.
\newblock \href {https://doi.org/10.18653/v1/2020.acl-main.173} {On
  faithfulness and factuality in abstractive summarization}.
\newblock In \emph{Proceedings of the 58th Annual Meeting of the Association
  for Computational Linguistics}, pages 1906--1919, Online. Association for
  Computational Linguistics.

\bibitem[{McKiernan(2000)}]{mckiernan2000arxiv}
Gerry McKiernan. 2000.
\newblock arxiv. org: the los alamos national laboratory e-print server.
\newblock \emph{International Journal on Grey Literature}.

\bibitem[{Nogueira et~al.(2020)Nogueira, Jiang, Cho, and
  Lin}]{nogueira2020navigationbased}
Rodrigo Nogueira, Zhiying Jiang, Kyunghyun Cho, and Jimmy Lin. 2020.
\newblock \href {http://arxiv.org/abs/2001.08687} {Navigation-based candidate
  expansion and pretrained language models for citation recommendation}.

\bibitem[{Papineni et~al.(2002)Papineni, Roukos, Ward, and
  Zhu}]{papineni-etal-2002-bleu}
Kishore Papineni, Salim Roukos, Todd Ward, and Wei-Jing Zhu. 2002.
\newblock \href {https://doi.org/10.3115/1073083.1073135} {{B}leu: a method for
  automatic evaluation of machine translation}.
\newblock In \emph{Proceedings of the 40th Annual Meeting of the Association
  for Computational Linguistics}, pages 311--318, Philadelphia, Pennsylvania,
  USA. Association for Computational Linguistics.

\bibitem[{Pride and Knoth(2020)}]{10.1145/3383583.3398617}
David Pride and Petr Knoth. 2020.
\newblock \href {https://doi.org/10.1145/3383583.3398617} {An authoritative
  approach to citation classification}.
\newblock In \emph{Proceedings of the ACM/IEEE Joint Conference on Digital
  Libraries in 2020}, JCDL '20, page 337–340, New York, NY, USA. Association
  for Computing Machinery.

\bibitem[{Radev et~al.(2013)Radev, Muthukrishnan, Qazvinian, and
  Abu-Jbara}]{10.1007/s10579-012-9211-2}
Dragomir~R. Radev, Pradeep Muthukrishnan, Vahed Qazvinian, and Amjad Abu-Jbara.
  2013.
\newblock \href {https://doi.org/10.1007/s10579-012-9211-2} {The acl anthology
  network corpus}.
\newblock \emph{Lang. Resour. Eval.}, 47(4):919–944.

\bibitem[{Radford et~al.(2019)Radford, Wu, Child, Luan, Amodei, and
  Sutskever}]{radford2019language}
Alec Radford, Jeff Wu, Rewon Child, David Luan, Dario Amodei, and Ilya
  Sutskever. 2019.
\newblock \href {https://openai.com/blog/better-language-models/} {Language
  models are unsupervised multitask learners}.

\bibitem[{Raffel et~al.(2020)Raffel, Shazeer, Roberts, Lee, Narang, Matena,
  Zhou, Li, and Liu}]{JMLR:v21:20-074}
Colin Raffel, Noam Shazeer, Adam Roberts, Katherine Lee, Sharan Narang, Michael
  Matena, Yanqi Zhou, Wei Li, and Peter~J. Liu. 2020.
\newblock \href {http://jmlr.org/papers/v21/20-074.html} {Exploring the limits
  of transfer learning with a unified text-to-text transformer}.
\newblock \emph{Journal of Machine Learning Research}, 21(140):1--67.

\bibitem[{See et~al.(2017)See, Liu, and Manning}]{see-etal-2017-get}
Abigail See, Peter~J. Liu, and Christopher~D. Manning. 2017.
\newblock \href {https://doi.org/10.18653/v1/P17-1099} {Get to the point:
  Summarization with pointer-generator networks}.
\newblock In \emph{Proceedings of the 55th Annual Meeting of the Association
  for Computational Linguistics (Volume 1: Long Papers)}, pages 1073--1083,
  Vancouver, Canada. Association for Computational Linguistics.

\bibitem[{Sellam et~al.(2020)Sellam, Das, and Parikh}]{sellam-etal-2020-bleurt}
Thibault Sellam, Dipanjan Das, and Ankur Parikh. 2020.
\newblock \href {https://doi.org/10.18653/v1/2020.acl-main.704} {{BLEURT}:
  Learning robust metrics for text generation}.
\newblock In \emph{Proceedings of the 58th Annual Meeting of the Association
  for Computational Linguistics}, pages 7881--7892, Online. Association for
  Computational Linguistics.

\bibitem[{Small(2018)}]{SMALL2018461}
Henry Small. 2018.
\newblock \href {https://doi.org/https://doi.org/10.1016/j.joi.2018.03.007}
  {Characterizing highly cited method and non-method papers using citation
  contexts: The role of uncertainty}.
\newblock \emph{Journal of Informetrics}, 12(2):461--480.

\bibitem[{Xing et~al.(2020)Xing, Fan, and Wan}]{xing-etal-2020-automatic}
Xinyu Xing, Xiaosheng Fan, and Xiaojun Wan. 2020.
\newblock \href {https://doi.org/10.18653/v1/2020.acl-main.550} {Automatic
  generation of citation texts in scholarly papers: A pilot study}.
\newblock In \emph{Proceedings of the 58th Annual Meeting of the Association
  for Computational Linguistics}, pages 6181--6190, Online. Association for
  Computational Linguistics.

\end{thebibliography}
\bibliographystyle{acl_natbib}




\end{document}